\documentclass[conference]{IEEEtran}
\IEEEoverridecommandlockouts

\usepackage{cite}
\usepackage{amsmath,amssymb,amsfonts}
\usepackage{algorithmic}
\usepackage{graphicx}
\usepackage{textcomp}
\usepackage{xcolor}
\usepackage{multirow}
\usepackage{threeparttable}
\usepackage{hyperref}

\def\BibTeX{{\rm B\kern-.05em{\sc i\kern-.025em b}\kern-.08em
    T\kern-.1667em\lower.7ex\hbox{E}\kern-.125emX}}
\begin{document}

\title{A Quantitative Approach for Evaluating Disease Focus and Interpretability of Deep Learning Models for Alzheimer’s Disease Classification


\thanks{This work is supported by the NIH grant R01GM140467.}

}

\author{
\IEEEauthorblockN{Thomas Yu Chow Tam \textsuperscript{\dag}\thanks{\textsuperscript{\dag} These authors contributed equally to this work and should be considered co-first authors.}}
\IEEEauthorblockA{\textit{Dept. of Health Information Management} \\
\textit{University of Pittsburgh}\\
Pittsburgh, PA, USA \\
yuchowt@alumni.cmu.edu}
\and
\IEEEauthorblockN{Litian Liang \textsuperscript{\dag}}
\IEEEauthorblockA{\textit{Computational Biology Department} \\
\textit{Carnegie Mellon University}\\
Pittsburgh, PA, USA \\
lianglt8800@gmail.com}
\and
\IEEEauthorblockN{Ke Chen}
\IEEEauthorblockA{\textit{School of Information Sciences} \\
\textit{University of Illinois Urbana-Champaign}\\
Urbana, USA \\
kec10@illinois.edu}
\and
\IEEEauthorblockN{Haohan Wang \textsuperscript{*}}
\IEEEauthorblockA{\textit{School of Information Sciences} \\
\textit{University of Illinois at Urbana-Champaign}\\
Champaign, IL, USA \\
haohanw@illinois.edu}
\and
\IEEEauthorblockN{Wei Wu \textsuperscript{*}\thanks{\textsuperscript{*} Corresponding Authors.}}
\IEEEauthorblockA{\textit{Computational Biology Department} \\
\textit{Carnegie Mellon University}\\
Pittsburgh, PA, USA \\
weiwu2@cs.cmu.edu}
}

\maketitle
\begin{abstract}

Deep learning (DL) models have shown significant potential in Alzheimer’s Disease (AD) classification. However, understanding and interpreting these models remains challenging, which hinders the adoption of these models in clinical practice.  Techniques such as saliency maps have been proven effective in providing visual and empirical clues about how these models work, but there still remains a gap in understanding which specific brain regions DL models focus on and whether these brain regions are pathologically associated with AD.

To bridge such gap, in this study, we developed a quantitative disease-focusing strategy to first enhance the interpretability of DL models using saliency maps and brain segmentations; then we propose a disease-focus (DF) score that quantifies how much a DL model focuses on brain areas relevant to AD pathology based on clinically known MRI-based pathological regions of AD. Using this strategy, we compared several state-of-the-art DL models, including a baseline 3D ResNet model, a pretrained MedicalNet model, and a MedicalNet with data augmentation to classify patients with AD vs. cognitive normal patients using MRI data; then we evaluated these models in terms of their abilities to focus on disease-relevant regions. Our results show interesting disease-focusing patterns with different models, particularly characteristic patterns with the pretrained models and data augmentation, and also provide insight into their classification performance. These results suggest that the approach we developed for quantitatively assessing the abilities of DL models to focus on disease-relevant regions may help improve interpretability of these models for AD classification and facilitate their adoption for AD diagnosis in clinical practice. The code is publicly available at \url{https://github.com/Liang-lt/ADNI}.


\end{abstract}


\section{Introduction}
According to the Alzheimer's Association, 
an estimated 6.7 million Americans age 65 and older are living with Alzheimer's disease (AD) in 2023 \cite{better2024alzheimer}, 
and thus, accurate diagnosis of AD is a paramount task.
Magnetic resonance imaging (MRI) is a crucial tool for identifying biomarkers associated with AD in both clinical and research settings \cite{frisoni2010clinical}.
MRI images are often used for differential diagnosis to rule out other potential causes of dementia, as well as for exploration of the brain structures of healthy aging populations and those of AD patients. Also, there is a growing interest in using MRI for early detection of AD \cite{frisoni2010clinical, johnson2012brain}.

AD patients often show atrophy of specific brain regions on MRI images \cite{johnson2012brain}. 
This atrophy is a key marker for the disease, and is often used in the diagnosis and assessment of AD. In particular, atrophy in the medial temporal lobe, including the hippocampus, entorhinal cortex, and amygdala, as well as ventricular enlargement correlate strongly with cognitive decline in AD, supporting their validity as biomarkers for AD diagnosis and progression \cite{frisoni2010clinical}. 

With the increasing predictive capability of deep learning (DL) algorithms and the widely available medical images like structural MRIs from large-scale multi-site studies such as the Alzheimer's Disease Neuroimaging Initiative (ADNI), DL models have been trained with these data 
and are utilized on various tasks such as classification of AD vs. cognitively normal (CN) patients as well as AD progression prediction \cite{wen2020convolutional} 
Among these DL methods, convolutional neural networks (CNNs) are widely used for image classification tasks due to their ability to capture spatial hierarchies in images. In the context of AD, CNNs are applied to 2D slices or 3D volumes of MRI scans to detect patterns of atrophy and other abnormalities associated with the disease. These methods achieve impressive performance with high accuracy on various AD classification tasks such as distinguishing AD vs. CN subjects and distinguishing AD, mild cognitive impairment (MCI), and CN subjects \cite{wen2020convolutional}.

However, one of the main obstacles that hinders the adoption of the DL models in clinical settings is their lack of interpretability and explainability. Despite the superior performance of the CNN-based models, 
these deep neural network models are often considered "black boxes" that make decisions based on complex, non-linear interactions within the data, not easily interpretable by humans. In a clinical setting, it is crucial for healthcare providers to understand why a model makes a particular diagnosis or prediction. 

To overcome such limitations, saliency maps have been employed to empirically highlight the areas of an input image that have the most significant impact on the output prediction of a DL model \cite{simonyan2013deep, adebayo2018sanity, arun2020assessing}. In particular, gradient-based saliency maps are a commonly used technique that is typically generated by calculating the gradient of the output prediction with respect to the input image. The gradients indicate how much a small change in each voxel would affect the prediction, with larger gradients suggesting that the corresponding voxels are more important to the decision. Despite this effort, there remains a gap in understanding which specific brain
regions the DL model focuses on and whether these brain regions are pathologically associated with AD.



In order to address this issue, in this study, we developed a strategy that combines saliency maps and brain segmentations to enhance the interpretability of DL models for AD classification; then we propose a disease-focus (DF) score that quantifies how much a DL model focuses on brain areas relevant to AD pathology based on clinically known MRI-based markers of AD. Using this strategy, we compared several state-of-the-art CNN-based DL models, including a 3D ResNet model \cite{hara2017learning}, a fine-tuned MedicalNet \cite{chen2019med3d} and the MedicalNet model with data augmentation techniques, as well as the volumetric feature-based machine learning (ML) methods using MRI scans in the Alzheimer’s Disease Neuroimaging Initiative (ADNI) database (adni.loni.usc.edu)
to assess the abilities of the DL models to focus on the pathologically important brain regions relevant to AD. Our results indicate that fine-tuning a pretrained model significantly enhances the model’s ability to concentrate on disease-relevant regions, while data augmentation further improves the model’s generalization capabilities by enabling it to learn features invariant in training samples, particularly those in the background. This approach for quantitatively assessing the disease-focusing areas of the DL models shows promise in improving their interpretability for AD classification and can facilitate their adoption for clinical practice.


\section{Methods}
In this work, our goal is to develop a strategy that can be used to compare DL models in terms of their abilities to focus on disease-relevant areas. Toward this goal, we consider a binary classification task that is to distinguish AD vs. CN subjects. In particular, participants are considered as AD if they were diagnosed as AD at baseline and stayed stable during the follow-up, whereas CN participants are those who were diagnosed as CN at baseline and stayed stable during the follow-up. These labels are available in the ADNI dataset.

\subsection{Convolutional Neural Networks with MRI Data}

We employed the CNN-based models to analyze 3D MRI data for Alzheimer's Disease classification. We mainly focus on three models: 3D ResNet, MedicalNet, 
and MedicalNet with Data Augmentation (MedicalNet + DA). 

\textbf{3D ResNet}: The baseline model used in our study was a 3D ResNet with 10 layers, also referred to as 3D ResNet-10 \cite{hara2017learning}. A 3D ResNet model extends the traditional 2D ResNet architecture into three dimensions, which is a better fit for voxel-based MRI data than a 2D model. The 3D ResNet model uses 3D convolutions to process the entire 3D volume of data, enabling the network to learn spatial hierarchies of features across the 3D space of the MRI scans, and thus capturing complex anatomical and pathological patterns in MRI data. The 3D ResNet model was trained from scratch using the Binary Cross-Entropy (BCE) loss function.


\textbf{MedicalNet}: MedicalNet is based on the same 3D ResNet structure but pretrained on a large-scale medical imaging dataset, known as the Medical Decathlon, which includes data from ten different medical imaging modalities covering a wide variety of diseases, such as brain tumors, liver tumors, and cardiac conditions \cite{chen2019med3d}. This diverse pretraining allows MedicalNet to leverage the learned features from these datasets, enabling it to generalize well across different types of medical images. In our experiments, we fine-tuned MedicalNet with the ADNI MRI data using the same BCE loss function.

\textbf{MedicalNet + DA}: To further enhance the performance of MedicalNet, we employed a series of data augmentation techniques during the training process. Data augmentation artificially increases the size and variability of the training dataset \cite{hussain2017differential}, helping to prevent overfitting and improve the model's ability to generalize to unseen data \cite{perez2017effectiveness, shorten2019survey}. The data augmentation techniques we used include random crop and random rotation, grey dilation, and grey erosion. The last two methods are morphological operations applied to the intensity values of the MRI scans, which can be particularly useful when dealing with grayscale images like medical scans. Specifically, Grey Dilation increases the brightness of regions by expanding the intensity values of the brightest voxels in a neighborhood, effectively simulating the swelling or expansion of tissues in the brain. Conversely, Grey Erosion decreases the brightness by reducing the intensity values of the brightest voxels, simulating the shrinking or erosion of tissues. These augmentations may help the model become robust to variations in tissue density and structural integrity, which are common in pathological brain conditions like AD. 

\subsection{Region-of-Interest (ROI) Segmentation and Volumetric Measurements from FastSurfer}

FastSurfer is a deep learning-based tool for the automated segmentation of brain MRI scans \cite{henschel2020fastsurfer}. 
FastSurfer allows for the segmentation of the ROIs that are pathologically important to AD, such as the hippocampus, entorhinal cortex, temporal lobes, ventricles, and cortical regions. 

FastSurfer can also provide volumetric measurements for all the ROIs, which include various cortical regions such as the frontal, temporal, parietal, and occipital lobes; subcortical structures such as the hippocampus, amygdala, thalamus, and the main ventricles.  

\subsection{Our Quantitative Disease-Focusing Approach}

In order to systematically quantify and compare the regions of the brain that the CNN models focus on when making predictions, we developed a two-step strategy to objectively assess which brain regions are prioritized by the CNN models during the prediction process, as well as how much these regions are pathologically associated with AD.

\textbf{Step I: Enhancing interpretability of the CNN models with saliency maps and brain segmentations.} This step allows us to visualize and measure a model’s attention to different brain areas within the input image. Specifically, we generated the saliency map for each model by computing the magnitude of the gradient of the predicted class score with respect to each input voxel.
The magnitude of these gradients indicates how much a small change in each voxel would affect the model's output, thus providing a visual map of the regions that the model considers important for its decision.
To achieve this, given a model’s output $y$ and an input image $X$, the saliency map $S$ is computed as:
\begin{equation}
S = |\frac{\partial y}{\partial X}|
\end{equation}
where $\frac{\partial y}{\partial X}$ represents the gradient of the output with respect to the input, as described in \cite{simonyan2013deep}. %

To ensure that the saliency maps are comparable across different subjects and models, we implemented a min-max normalization on each saliency map. Then we aggregated the normalized gradient values within each brain region as defined by the FastSurfer segmentations. For each brain region r, we calculated the median gradient intensity:
\begin{equation}
M_r = \text{median}\left( S_r \right)
\end{equation}
where $S_r$ represents the normalized gradient values within region $r$.
The median value was chosen as it is less sensitive to outliers and provides a robust measure of the overall importance of each region as perceived by the model.

This region-wise analysis of the saliency maps allowed us to systematically compare the model's focus on different anatomical regions of the brain. By examining which brain regions consistently exhibited higher median gradient intensities, we can infer which areas the model relies on most heavily for making predictions. 

This step allows us to identify the top 10 brain regions with the highest median gradient intensities for each model.

\textbf{Step II: Assessing DL models using Disease-Focus Scores}
In order to quantify how much a DL model focuses on pathologically important brain areas relevant to AD, we developed a scoring system, termed the \textbf{Disease-Focus Score}, by leveraging clinical knowledge on known MRI-based biomarkers of AD as well as by literature survey.

In particular, in this scoring system, we first identified three categories (C1-C3) of brain regions showing differential pathological importance to AD: 

\textbf{C1:} This category contains the most important regions with severe change in AD, including hippocampus, entorhinal cortex, amygdala, medial temporal lobe (MTL), areas within MTL, as well as ventricular enlargement, as described in \cite{braak1991neuropathological, frisoni2010clinical}.

\textbf{C2:} This category includes the regions showing changes in AD, but are not as critically important as those in C1, e.g., supramarginal cortex, precentral cortex, pallidum, and pars opercularis, as described in \cite{yang2019study, van2023subcortical, mascali2018disruption} 

\textbf{C3:} This category consists of the rest of the ROIs in the brain showing no pathological changes in AD. 

Next, we assigned the Disease-Focus Score of 0, 1, or 2 to each brain region based on the category it belongs to, with 2 assigned to C1 which has the most pathologically important regions, and 0 to C3 which has the regions unaffected by AD. 

Finally, we calculated the average disease-focus (DF) score for the top 10 brain regions identified for each model from Step I. A higher DF score indicates that a model can more effectively identify and focus on the pathological regions associated with Alzheimer's Disease.

\begin{equation}
\text{DF Score} = \text{Average} (\text{Score}(r_i))
\end{equation}
where $\text{Score}(r_i)$ represents the Disease-Focus Score assigned to the $i$-th region in the top 10 list.


This integrated analysis provides a quantitative framework for evaluating the alignment of the DL model's focus areas with pathological reality, offering a deeper understanding of the model's decision-making processes relevant to AD.

\section{Experiments}

\subsection{Experimental Setup}
\textbf{Datasets:} 3D T1-weighted structural MRI scans from the ADNI dataset were used to train the selected DL and ML models in this study. Proper preprocessing procedures including skull stripping, B1 bias field correction, and normalization were performed with FastSurfer \cite{henschel2020fastsurfer}, following the preprocessing steps proposed in \cite{wen2020convolutional}. The output MRIs from the preprocessing pipeline were then used in subsequent model training. The images were split into training and test sets based on participant IDs to ensure there was no data leakage. 
In our training set, there are 456 participants in total  and 1,675 images, of which 839 are from AD subjects and 836 from CN subjects); 
in the internal ADNI test set, there are 93 participants in total and 389 images, of which 137 are from AD subjects and 252 from CN subjects).




After training the selected DL and ML models, we performed an independent test of the models using 3D T1-weighted structural MRI scans from the Australian Imaging, Biomarkers \& Lifestyle Flagship Study of Ageing (AIBL) dataset (see www.aibl.csiro.au for further details). 
The independent test set consists of 100 participants in AIBL, and 360 images of which 180 are from AD subjects and 180 from CN subjects.

\textbf{Data Preprocessing:} The MRI scans from both the ADNI and AIBL datasets were preprocessed with FastSurfer using standard protocols including skull stripping, B1 bias field correction, and, normalization as proposed in \cite{wen2020convolutional}. 

Brain segmentation maps were generated using the FastSurfer, which provided volumetric features for various brain regions.



\textbf{Model Architectures and Training:} We compared several deep learning methods in this study, including three CNN models with MRI data and two CNN models with ROI segmentations from FastSurfer. For comparison purposes, we also included 4 conventional volumetric feature-based ML (VF-ML) approaches using ROI segmentation statistics as reference methods.







\subsubsection{CNNs with MRI Data} In our experiments, we employed three primary CNN models for Alzheimer’s Disease classification using MRI data: the 3D ResNet model trained from scratch, MedicalNet, and MedicalNet with data augmentation. Additionally, we explored the use of ROI segmentation data generated by FastSurfer as input for training two CNN models, the 3D ResNet from scratch and MedicalNet.



\subsubsection{Volumetric feature-based ML Approach with ROI Segmentation Statistics}
As reference methods for comparing the CNN models,
we utilized four different VF-ML algorithms: Support Vector Machine (SVM) with a Radial Basis Function (RBF) kernel, Random Forest, Extreme Gradient Boosting (XGBoost), and Logistic Regression with L2 regularization. These models were trained using the 700 volumetric features extracted from the ROI segmentation statistics generated by FastSurfer.

\textbf{Evaluation metrics:} To evaluate the performance of the compared models in terms of their abilities to accurately classify Alzheimer's Disease, we computed several metrics including specificity, sensitivity, F-1 score (F-1), Balanced Accuracy (BA), Area Under the Receiver Operating Characteristic Curve (AUROC), and Area Under the Precision-Recall Curve (AUPRC). 

\textbf{Implementation details:} For training and testing the models, we utilized Pittsburgh Supercomputing Center servers Bridges-2 GPU with 8 NVIDIA Tesla V100-32GB SXM2 GPUs and storage Ocean. The implementation of the experiments was performed with Python (version 3.10.12) and PyTorch (version 2.0.1). Our DL models were trained for 100 epochs with a batch size of 6; Cross-entropy loss and optimizer Adam with a learning rate of 1e-4 were used. 
All the machine learning models were constructed with scikit-learn and using default settings. 

\subsection{Independent Test with an External Dataset}

To assess the generalization capability of the trained models, we conducted independent testing using the AIBL dataset. The same evaluation metrics and implementation details were used to compare the performance of the models on this external dataset.



\subsection{Comparisons of the CNN models using Our Quantitative Disease-Focusing Approach}

\subsubsection{Saliency Map Generation}

Saliency maps were generated using a gradient-based method as described in the Methods section and implemented using PyTorch.
Specifically, for each input MRI scan, we computed the gradient of the predicted class score with respect to the input image voxels. 
Saliency maps were generated for all test images.

\subsubsection{Analysis of Saliency Maps}


We performed a detailed analysis to quantify a model's attention to different brain regions using our disease-focusing approach as described in Methods. Specifically, for each brain region, we first computed the median gradient intensity $M_r$ across all voxels within that region, and then based on the magnitude of $M_r$, we identified the top 10 brain regions for each CNN model. 

Next, we calculated the average Disease-Focus Score using the top 10 brain regions identified by each model for each test image. Models that achieved higher DF scores are considered to have a better capability of focusing on pathological regions relevant to Alzheimer's disease.


\subsubsection{Evaluation of the disease focusing ability of the VF-ML methods}
For comparison purposes, we also assess the disease-focusing ability of the VF-ML methods using mutual information. Mutual information measures the nonlinear dependency of two variables and thus allows us to measure the association of AD and each volumetric feature corresponding to specific brain regions. 

\section{Results}

\begin{table*}[t]
\centering
\caption{Performance on ADNI Test Set}
\label{table:dl_results}
\begin{tabular}{lllllllll}
\hline
\textbf{Model} & \textbf{Data} & \textbf{F-1} & \textbf{BA} & \textbf{AUROC} & \textbf{AUPRC} & \textbf{Specificity} & \textbf{Sensitivity} \\
\hline
\hline
3D ResNet & MRI & 0.86 & 0.86 & 0.91 & 0.89 & 0.87 & 0.85 \\ 
MedicalNet & MRI & 0.87 & 0.86 & 0.93 & 0.88 & \textbf{0.90} & 0.82 \\
MedicalNet + DA & MRI & \textbf{0.89} & \textbf{0.89} & \textbf{0.94} & \textbf{0.93} & 0.88 & \textbf{0.91} \\
\hline
3D ResNet & Segmentation & 0.80 & 0.79 & 0.85 & 0.80 & 0.84 & 0.73 \\
MedicalNet & Segmentation & 0.79 & 0.80 & 0.91 & 0.90 & 0.77 & 0.80 \\
\hline
SVM & Volumetric Features & \textbf{0.93} & \textbf{0.93} & \textbf{0.98} & \textbf{0.97} & \textbf{0.96} & \textbf{0.88} \\
XGBoost & Volumetric Features & 0.92 & 0.91 & 0.97 & 0.96 & 0.95 & 0.88 \\
Random Forest& Volumetric Features & 0.92 & 0.90 & 0.97 & 0.95 & 0.95 & 0.85 \\
Logistic Regression & Volumetric Features & 0.88 & 0.87 & 0.93 & 0.92 & 0.91 & 0.82 \\
\hline
\end{tabular}
\end{table*}


\begin{table*}[t]
\centering
\caption{Independent Test on AIBL dataset}
\label{table:ind_results}
\begin{tabular}{llllllll}
\hline
\textbf{Model} & \textbf{Data} & \textbf{F-1} & \textbf{BA} & \textbf{AUROC} & \textbf{AUPRC} & \textbf{Specificity} & \textbf{Sensitivity} \\
\hline
\hline
3D ResNet & MRI & 0.89 & 0.80 & 0.89 & 0.67 & 0.92 & 0.68  \\
MedicalNet & MRI & 0.92& 0.74 & 0.91 & 0.68& \textbf{0.97} & 0.52 \\
MedicalNet + DA & MRI & \textbf{0.92} & \textbf{0.81} & \textbf{0.93} & \textbf{0.73} & 0.96 & 0.66  \\
\hline
3D ResNet & Segmentation & 0.88 & 0.74 & 0.86 & 0.49 & 0.93  & 0.56  \\
MedicalNet & Segmentation & 0.83 & 0.80 & 0.89 & 0.62 & 0.83 & \textbf{0.76} \\

\hline

SVM & Volumetric Features & 0.88 & 0.79 & 0.88 & 0.71 & 0.91& 0.66 \\
XGBoost & Volumetric Features & \textbf{0.93} & 0.80 & 0.89 & \textbf{0.76} & \textbf{0.98} & 0.62 \\
RandomForest & Volumetric Features & 0.92 & 0.82 & 0.90 & 0.72 & 0.96& 0.68  \\
Logistic Regression & Volumetric Features & 0.92 & \textbf{0.83} & \textbf{0.91} & 0.70 & 0.95& \textbf{0.70}   \\
\hline






\end{tabular}
\end{table*}

\subsection{Classification Performance}

First, we evaluated the classification performance of the CNN models and the VF-ML models for AD diagnosis using the ADNI and AIBL datasets. Table \ref{table:dl_results} presents the results on the ADNI test set.
Our results show that the fine-tuned MedicalNet model with data augmentation outperforms other models on most metrics, highlighting the effectiveness of pretraining and data augmentation in improving the classification performance of the model. 

We also investigated how brain segmentation data influences AD classification. When we trained the 3D ResNet using ROI segmentations from FastSurfer, the test performance of the model generally lags behind the CNN models using MRIs. Furthermore, applying pretrained weights from MedicalNet and finetuning with brain segmentation data does not improve the performance.

Table \ref{table:dl_results} also shows the results for the VF-ML models using volumetric features. Almost all the VF-ML models (except logistics regression) outperform the best DL model MedicalNet + DA. In particular, SVM scores the highest across all models in all evaluation metrics, notably, the AUROC and AUPRC score 0.98 and 0.97, respectively, close to the theoretical limit of 1.

The results of the independent test obtained on the AIBL dataset, shown in Table \ref{table:ind_results}, provide insight into the generalization ability of the models on unseen data. For the DL models, the analysis reveals that while the 3D ResNet and MedicalNet models both perform well on MRI data, with F-1 scores of 0.89 and 0.92, respectively, there are notable differences in other metrics. Adding data augmentation to MedicalNet improves the balance between sensitivity and specificity, with an increase in Balanced Accuracy (BA) to 0.81 and an AUROC of 0.93. The AIBL dataset, often referred to as the Australian ADNI, is closely aligned with the ADNI dataset, as both share a collaborative infrastructure supported by the Alzheimer’s Association. The similarity in data collection and goals between AIBL and ADNI means that models trained on the ADNI dataset are well-suited for evaluation on AIBL data. This alignment ensures that models trained on ADNI continue to perform well during independent tests on AIBL data, highlighting their robust generalization capabilities. The ML models trained with volumetric features continue to perform well, with XGBoost achieving the highest F-1 score of 0.93. This demonstrates that the conventional VF-ML approaches remain robust and effective.

However, when using brain segmentation data, both 3D ResNet and MedicalNet models show a decrease in performance, particularly in Sensitivity. This suggests that segmentation data, while informative, may not be as effective as MRI data alone in this context. Furthermore, combining MRI and segmentation data does not consistently improve performance, which is consistent with the trends observed in the ADNI dataset, where models trained on MRI data generally outperformed those using segmentation data alone.


\subsection{Saliency Map Analysis}

\begin{figure*}[t]
\centering
\includegraphics[width=0.9\textwidth]{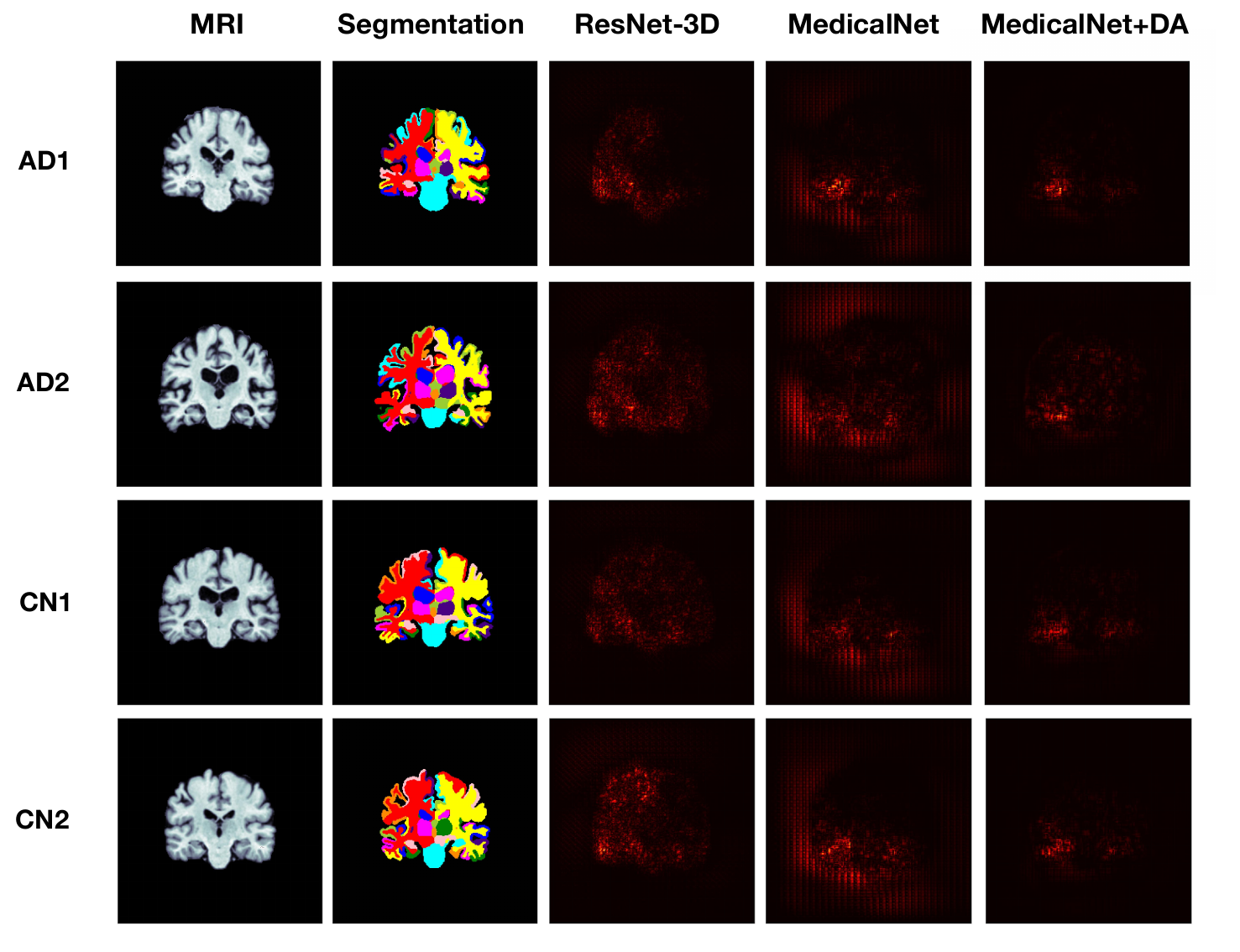}
\caption{Saliency maps illustrating the areas of an input image with the largest effect on the output prediction of DL models.
Saliency maps for 2 AD subjects and 2 CN subjects are shown in the rows. The columns show the MRI scans, ROI-segmented images, and the saliency maps for the 3D ResNet, MedicalNet, and MedicalNet + DA, respectively.
}
\label{fig:saliency_maps}
\end{figure*}

\begin{table*}[t]
\centering
\caption{Disease-focusing Analysis for Different CNN Models}
\label{table:saliency_map_analysis}
\begin{threeparttable}
\begin{tabular}{lllll}
\hline
\textbf{Model} & \multicolumn{2}{c}{\textbf{3D ResNet}} & \multicolumn{2}{c}{\textbf{MedicalNet / MedicalNet + DA}} \\ 
\hline
\textbf{Classification} & \textbf{True Positive} & \textbf{True Negative} & \textbf{True Positive} & \textbf{True Negative} \\ 
\hline
\multirow{10}{*}{\textbf{Regions}} & \textcolor{red}{ctx-rh-entorhinal} & \textcolor{red}{ctx-rh-entorhinal} & \textcolor{red}{Left-Inf-Lat-Vent} & \textcolor{red}{Left-Inf-Lat-Vent} \\ 
 & \textcolor{red}{Left-Inf-Lat-Vent} & \textcolor{red}{Left-Inf-Lat-Vent} & \textcolor{red}{ctx-rh-entorhinal} & \textcolor{red}{ctx-rh-entorhinal} \\ 
 & \textcolor{red}{Left-Amygdala} & \textcolor{red}{Left-Amygdala} & \textcolor{red}{ctx-lh-entorhinal} & \textcolor{red}{ctx-lh-parahippocampal} \\ 
 & \textcolor{red}{Left-Hippocampus} & \textcolor{red}{Left-Hippocampus} & \textcolor{red}{Left-Hippocampus} & \textcolor{red}{Left-Hippocampus} \\ 
 & \textcolor{red}{Right-Amygdala} & \textcolor{red}{Right-Amygdala} & \textcolor{red}{ctx-lh-parahippocampal} & \textcolor{red}{ctx-lh-entorhinal} \\ 
 & \textcolor{blue}{Left-Pallidum} & \textcolor{blue}{Left-Pallidum} & \textcolor{red}{Right-Inf-Lat-Vent} & \textcolor{red}{Right-Inf-Lat-Vent} \\ 
 & \textcolor{red}{Right-Inf-Lat-Vent} & \textcolor{red}{Right-Inf-Lat-Vent} & \textcolor{red}{ctx-lh-inferiortemporal} & \textcolor{red}{ctx-lh-inferiortemporal} \\ 
 & \textcolor{red}{ctx-lh-entorhinal} & \textcolor{blue}{ctx-lh-supramarginal} & \textcolor{red}{Right-Amygdala} & \textcolor{red}{Right-Amygdala} \\ 
 & \textcolor{blue}{ctx-lh-parsopercularis} & \textcolor{red}{ctx-lh-middletemporal} & \textcolor{red}{Left-Amygdala} & \textcolor{red}{Left-Amygdala} \\ 
 & \textcolor{red}{ctx-lh-middletemporal} & \textcolor{blue}{ctx-lh-precentral} & \textcolor{blue}{Left-Pallidum} & \textcolor{blue}{Left-Pallidum} \\ 
\hline
\textbf{DF Score} & \multicolumn{2}{c}{\textbf{1.75}}  & \multicolumn{2}{c}{\textbf{1.9}}   \\ 

\hline

\end{tabular}
    \begin{tablenotes}    
        \footnotesize    
                 \item[] Notes: Brain regions shown in \textcolor{red}{red} are in the C1 category; those in \textcolor{blue}{blue} are in the C2 category.

         \item[] \textbf{Abbreviations}: ctx, cortex;  Inf-Lat-Vent, inferior lateral ventricles; lh, left hemisphere; rh, right hemisphere.
      \end{tablenotes}
\end{threeparttable}      
\end{table*}

\begin{table*}[t]
\begin{center}
\begin{threeparttable}

\caption{Saliency Map Analysis for MedicalNet + DA, categorized by AD classification results 
}
\label{table:FP_analysis}
\begin{tabular}{llllllll}
\hline
\textbf{True Positive} & \textbf{$M_r$} & \textbf{True Negative} & \textbf{$M_r$} & \textbf{False Positive} & \textbf{$M_r$} & \textbf{False Negative} & \textbf{$M_r$} \\
\hline
Left-Inf-Lat-Vent & 0.386 & Left-Inf-Lat-Vent & 0.195 & Left-Inf-Lat-Vent & 0.080 & Left-Inf-Lat-Vent & 0.094 \\
ctx-rh-entorhinal & 0.288 & ctx-rh-entorhinal & 0.119 & ctx-rh-entorhinal & 0.062 & ctx-lh-parahippocampal & 0.056 \\
ctx-lh-entorhinal & 0.236 & ctx-lh-parahippocampal & 0.108 & Left-Hippocampus & 0.050 & ctx-rh-entorhinal & 0.052 \\
Left-Hippocampus & 0.229 & Left-Hippocampus & 0.104 & ctx-lh-parahippocampal & 0.049 & Left-Hippocampus & 0.051 \\
ctx-lh-parahippocampal & 0.222 & ctx-lh-entorhinal & 0.100 & ctx-lh-entorhinal & 0.046 & ctx-lh-inferiortemporal & 0.039 \\
Right-Inf-Lat-Vent & 0.206 & Right-Inf-Lat-Vent & 0.093 & Right-Inf-Lat-Vent & 0.045 & ctx-lh-entorhinal & 0.039 \\
ctx-lh-inferiortemporal & 0.200 & ctx-lh-inferiortemporal & 0.090 & Right-Amygdala & 0.039 & Right-Inf-Lat-Vent & 0.039 \\
Right-Amygdala & 0.198 & Right-Amygdala & 0.077 & ctx-lh-inferiortemporal & 0.038 & Left-Amygdala & 0.035 \\
Left-Amygdala & 0.180 & Left-Amygdala & 0.077 & Left-Amygdala & 0.036 & Right-Amygdala & 0.034 \\
Left-Pallidum & 0.156 & Left-Pallidum & 0.067 & Left-Pallidum & 0.031 & Left-Pallidum & 0.032 \\
\hline
\end{tabular}
\begin{tablenotes}    
    \footnotesize               
     \item[] Abbreviations: ctx, cortex;  Inf-Lat-Vent, inferior lateral ventricles; lh, left hemisphere; rh, right hemisphere.
\end{tablenotes}

\end{threeparttable}
\end{center}
\end{table*}

In order to gain an empirical understanding of the interpretability of the CNN models, we examined the saliency maps for each test image illustrating how each voxel in the image influences the prediction output for the three CNN models, 3D ResNet, MedicalNet, and MedicalNet + DA.
Figure \ref{fig:saliency_maps} shows the saliency maps for two AD and two CN subjects. 
For the 3D ResNet model, the saliency maps demonstrate a rather dispersed pattern of attention. The model tends to distribute its focus across large, non-specific areas of the brain rather than concentrating on pathological regions typically associated with AD, suggesting 3D ResNet’s limited ability to detect pathologically important areas. 

In contrast, it is noticeable that the MedicalNet model exhibits significant attention to the background areas. This unexpected focus on areas other than the brain can be attributed to the model’s pretraining on a diverse set of medical imaging data. During the pretraining phase, the 
model may have learned to focus on features relevant to a broader set of medical images, some of which include different scales, resolutions, and preprocessing techniques.

Since the MRI images in the ADNI dataset typically have consistent background regions (e.g., black areas outside the brain), the lack of variability in these regions during fine-tuning may lead the model to maintain its pretrained focus on irrelevant areas like the background. However, even with this issue, MedicalNet demonstrates significantly more concentrated attention on key pathological areas, such as the hippocampus, entorhinal cortex, and amygdala, compared to 3D ResNet. These regions are crucial for memory and cognitive function, and their atrophy is strongly associated with Alzheimer’s Disease.

For the MedicalNet + DA model, this background-attention issue is notably resolved. The data augmentation techniques applied—random rotation, random cropping, grey dilation, and grey erosion—not only increase the variability of the training data, but also inadvertently ensure that the model learns to discount irrelevant background information. This adjustment allows the model to maintain a focused attention on critical brain regions, particularly within the medial temporal lobe, which includes the hippocampus, entorhinal cortex, and amygdala. These areas are crucial for understanding Alzheimer’s Disease, and the model’s ability to consistently highlight them, while reducing background focus, likely contributes to its overall performance, as reflected in the saliency maps (Figure \ref{fig:saliency_maps}) and the statistical results presented in our tables.




In summary, the fine-tuned model MedicalNet model significantly improves its ability to concentrate attention on key pathological regions of the brain associated with Alzheimer’s Disease, as evidenced by the more focused saliency maps. However, while data augmentation does not further enhance the model’s focus on these critical regions—as reflected in the rank in Table \ref{table:saliency_map_analysis}—it does lead to a notable improvement in overall classification performance. This improvement is likely due to the model’s ability to learn features that are invariant to small perturbations in the data, thus enhancing its generalization capabilities. The consistent alignment between the refined attention patterns from fine-tuning and the improved performance metrics underscores the value of interpretability in evaluating model effectiveness. Meanwhile, the contribution of data augmentation to robustness, even without directly improving focus on disease-relevant regions, highlights its role in making MedicalNet the most powerful model in our study.

\subsection{Disease-Focusing Quantitative Analysis}

Next, we quantitatively analyzed the saliency maps generated for the 3 CNN models using our disease-focusing approach.
Table \ref{table:saliency_map_analysis} presents the top 10 brain regions for each model, categorized by true positive (TP) and true negative (TN) samples based on their AD classification with MRI data. 

The analysis reveals that the 3D ResNet model shows significant discrepancies between the rankings for TP and TN, suggesting its instability in feature learning; whereas for MedicalNet and MedicalNet + DA, the rankings for TP and TN are almost identical, indicating that these pretrained models have consistent feature learning abilities. Additionally, comparing the models shows that the ranking order for the 3D ResNet model differs significantly from the other two pretrained models. This discrepancy likely arises from the non-pretrained model’s weaker feature learning capabilities, while the pretrained models can better capture important features in the data. These findings demonstrate that pretraining significantly enhances the stability and accuracy of feature learning in classification tasks, thereby improving the models’ generalization capabilities.

To further quantify the clinical relevance of the identified regions, we applied the \textbf{Disease-Focus Score} system as we introduced in the Method section. This system calculates the average Disease-Focus Score for the top 10 regions identified by each model. The brain regions were grouped into three categories based on clinically known MRI-based pathological biomarkers in Alzheimer’s Disease: Category 1 (\textcolor{red}{red} in Table \ref{table:saliency_map_analysis}) includes regions with undisputed significance, such as the hippocampus, entorhinal cortex, amygdala, and inferior lateral ventricles; Category 2 (\textcolor{blue}{blue} in Table \ref{table:saliency_map_analysis}) contains regions with some influence but generally less importance, including the pallidum, precentral gyrus, the supramarginal gyrus and pars opercularis, which are involved in broader neurological functions but are less directly associated with early AD pathology. There is no brain region considered in Category 3 within the top 10 regions identified by the models.
Models achieving higher Disease-Focus Scores are considered to have a better understanding of AD pathology, as they focus more accurately on important disease-related regions. Our results show that both MedicalNet and MedicalNet + DA achieve the DF scores of 1.9, outperforming the 3D ResNet which scored 1.75. These results indicate that MedicalNet and MedicalNet + DA models not only perform better in terms of classification metrics, but also demonstrate a more effective identification of key pathological areas relevant to Alzheimer’s Disease, further validating their interpretability and clinical relevance.

From Table \ref{table:FP_analysis} showing the saliency map analysis results for MedicalNet + DA, we observe that the median gradient intensity $M_r$ for true positive (TP) and true negative (TN) samples is significantly higher than those for false positive (FP) and false negative (FN) samples. This phenomenon suggests that the CNN model is more confident in its correct predictions (TP and TN), where the median gradient magnitude of the top regions is significantly higher, indicating a stronger focus on these areas in the saliency maps. In contrast, the lower gradient intensities in the FP and FN samples may indicate that the model’s decision is less certain, leading to misclassifications. This disparity in gradient intensities can be attributed to the model’s ability to capture the distinguishing features of AD more effectively in correctly classified samples. For FP and FN samples, the model may focus on less relevant or noisier regions, resulting in lower gradient values.

Also, we ``interpreted" the performance of the VF-ML methods using mutual information.
Table \ref{table:top_features} lists the top 10 predictive features sorted by mutual information obtained from the volumetric data. These features identify several brain regions that are crucial in the context of AD, including the entorhinal cortex, hippocampus, and amygdala, among others.  
The Disease-Focus Score calculated in Table \ref{table:top_features}, which reaches a value of 2.0—the highest possible score—highlights the clinical relevance of these top-ranking features. It provides further evidence of why the examined VF-ML models which utilize these features as inputs perform exceptionally well.

\begin{table}[t]
\centering
\begin{threeparttable}
\caption{Top 10 Features Identified by Mutual Information}
\label{table:top_features}
\begin{tabular}{ll}
\hline
\multicolumn{2}{c}{\textbf{Mutual Information}} \\
\hline
\textbf{Features} & \textbf{Regions} \\
\hline
Left-Lateral-Ventricle\_normStdDev & \textcolor{red}{Lateral-Ventricle (L)} \\
Left-Inf-Lat-Vent\_normMean & \textcolor{red}{Inf-Lat-Vent (L)} \\
Left-Hippocampus\_NVoxels & \textcolor{red}{Hippocampus (L)} \\
Left-Hippocampus\_Volume\_mm3 & \textcolor{red}{Hippocampus (L)} \\
Left-Amygdala\_NVoxels & \textcolor{red}{Amygdala (L)} \\
Left-Amygdala\_Volume\_mm3 & \textcolor{red}{Amygdala (L)} \\
Right-Inf-Lat-Vent\_normMean & \textcolor{red}{Inf-Lat-Vent (R)} \\
Right-Hippocampus\_NVoxels & \textcolor{red}{Hippocampus (R)} \\
Right-Hippocampus\_Volume\_mm3 & \textcolor{red}{Hippocampus (R)} \\
Right-Amygdala\_NVoxels & \textcolor{red}{Amygdala (R)} \\
\hline
\hline
\textbf{DF Score} & \textbf{2.0} \\
\hline
\end{tabular}
    \begin{tablenotes}    
         \footnotesize               
          \item[] Notes: Brain regions shown in \textcolor{red}{red} are in the C1 category. \item[] \textbf{Abbreviations}: Inf-Lat-Vent, inferior lateral ventricles; L, left hemisphere; R, right hemisphere. 
       \end{tablenotes}  
   \end{threeparttable}  
\end{table}

\section{Conclusion}

In summary, we developed a two-step quantitive disease-focusing approach which first combines saliency maps and brain segmentations to enhance the interpretability of the DL models for AD classification and then utilizes the Disease-Focus Score to quantify a DL model's attention on the brain areas pathologically important to AD. Using such a strategy, we performed an in-depth evaluation of several state-of-the-art CNN models for AD classification to compare their ability to focus on brain regions important for AD pathology. Our results provide insight into the classification performance of the models, particularly in light of how fine-tuning and data augmentation improve model performance. 


Our study also highlights the superior performance of conventional ML models using brain volumetric statistics over the CNN models for AD classification. Our results emphasize the need for improved interpretability in DL models to facilitate their clinical adoption. Integrating domain-specific clinical knowledge and refining saliency map techniques are crucial steps toward this goal.

\section*{Acknowledgment}
The authors wish to thank Yihan Zhang for preprocessing the ADNI MRI data.

This work used Bridges-2 GPU and Ocean at Pittsburgh Supercomputing Center through allocation CIS230151 from the Advanced Cyberinfrastructure Coordination Ecosystem: Services \& Support (ACCESS) program which is supported by National Science Foundation grants \#2138259, \#2138286, \#2138307, \#2137603, and \#2138296.

\bibliographystyle{IEEEtran}
\bibliography{reference}
\end{document}